%% file: main.tex
\documentclass{article}

\PassOptionsToPackage{round}{natbib}
\usepackage[preprint]{neurips_2024}

\usepackage[utf8]{inputenc} 
\usepackage[T1]{fontenc}    
\usepackage{hyperref}       
\usepackage{url}            
\usepackage{booktabs}       
\usepackage{amsfonts}       
\usepackage{nicefrac}       
\usepackage{microtype}      
\usepackage{xcolor}         
\usepackage{graphicx}
\usepackage{amsmath}
\usepackage{siunitx}
\usepackage{etoc} 
\usepackage{makecell}
\usepackage{bibunits}
\usepackage{enumitem}

\newcommand{\unete}{U-Net\textsubscript{DOW-6}}

\newcommand{\dinov}{DINOv2}

\newcommand{\uneth}{U-Net\textsubscript{DOW}}
\newcommand{\unetm}{U-Net\textsubscript{Multi}}
\newcommand{\unethm}{U-Net\textsubscript{DOW-Multi}}

\newcommand{\dinovs}{DINOv2\textsubscript{Multi}}
\newcommand{\dinovh}{DINOv2\textsubscript{DOW}}
\newcommand{\dinovsh}{DINOv2\textsubscript{DOW-Multi}}
\newcommand{\yolom}{YOLOv8\textsubscript{Multi}}
\newcommand{\yolob}{YOLOv8}

\newcommand{\apf}{AP\textsubscript{50}}
\newcommand{\mapf}{mAP\textsubscript{50}}
\newcommand{\apfn}{AP\textsubscript{50-95}}
\newcommand{\mapfn}{mAP\textsubscript{50-95}}
\newcommand{\tps}{TP\textsubscript{s}}
\newcommand{\iout}{mIoU\textsuperscript{3}}
\newcommand{\iouf}{mIoU\textsuperscript{5}}
\newcommand{\apft}{mAP\textsubscript{50}\textsuperscript{3}}
\newcommand{\apff}{mAP\textsubscript{50}\textsuperscript{5}}
\renewcommand{\apft}{mAP$_{\text{50}}^{\text{3}}$}
\renewcommand{\apff}{mAP$_{\text{50}}^{\text{5}}$}

\newcommand{\ds}{Nacala-Roof-Material}
\newcommand{\D}{\mathcal{D}}
\newcommand{\dtrain}{$\D_{\text{train}}$}
\newcommand{\dval}{$\D_{\text{val}}$}
\newcommand{\dtestone}{$\D_{\text{test}}$}
\newcommand{\dtesttwo}{$\D_{\text{ext}}$}

\newcommand{\ccell}[2]{\makecell{#1 \\[-2.5pt] \scriptsize $\pm$ #2}}
\newcommand{\bcell}[2]{\makecell{\textbf{#1} \\[-2.5pt] \scriptsize $\pm$ #2}}

\newcommand{\np}{n_{\text{pix}}}
\newcommand{\nl}{n_{\text{lev}}}
\newcommand{\ngap}{n_{\text{gap}}}

\title{\ds{}: Drone Imagery for Roof Detection, Classification, and Segmentation
to Support Mosquito-borne Disease Risk Assessment}

\author{%
Venkanna Babu Guthula\\University of Copenhagen, DK\\\texttt{vegu@di.ku.dk}\And
Stefan Oehmcke\\University of Rostock, DE\And
Remigio Chilaule\\Royal Danish Academy, DK\\\#MapeandoMeuBairro, MZ\And
Hui Zhang\\University of Copenhagen, DK\And
Nico Lang\\University of Copenhagen, DK\And 
Ankit Kariryaa\\University of Copenhagen, DK\And 
Johan Mottelson\\Royal Danish Academy, DK\And 
Christian Igel\\University of Copenhagen, DK
}

\begin{document}

\maketitle

\begin{abstract}
As low-quality housing and in particular certain roof characteristics are associated with an increased risk of malaria, classification of roof types based on remote sensing imagery can support the assessment of malaria risk and thereby help prevent the disease. To support research in this area, we release the Nacala-Roof-Material dataset, which contains high-resolution drone images from Mozambique with corresponding labels delineating houses and specifying their roof types.
The dataset defines a multi-task computer vision problem, comprising object detection, classification, and segmentation.
In addition, we benchmarked various state-of-the-art approaches on the dataset. Canonical U-Nets, YOLOv8, and a custom decoder on pretrained DINOv2 served as baselines. We show that each of the methods has its advantages but none is superior on all tasks, which highlights the potential of our dataset for future research in multi-task learning. While the tasks are closely related, accurate segmentation of objects does not necessarily imply accurate instance separation, and vice versa. 
We address this general issue by introducing a variant of the deep ordinal watershed (DOW) approach that additionally separates the interior of objects, allowing for improved object delineation and separation. 
We show that our DOW variant is a generic approach that improves the performance of both U-Net and DINOv2 backbones, leading to a better trade-off between semantic segmentation and instance segmentation.
\end{abstract}

\section{Introduction}
Mosquito-borne diseases refer to a group of infectious illnesses transmitted by the bite of mosquitoes. Malaria is a mosquito-borne disease caused by single-celled parasites of the Plasmodium group spread through bites of infected female Anopheles mosquitoes. It ranks among the world's most severe public health problems and is a leading cause of mortality and disease in many developing countries. It is therefore crucial to improve prevention, control, and surveillance measures of malaria, particularly in sub-Saharan Africa \citep{VENKATESAN2024e214,who2023}. 
Low-quality housing built of natural materials, for example, having a thatched roof of grass or palm and having cane, grass, shrub, or mud as internal and external walls, is associated with an increased risk of malaria infection \citep{nomcebo_2017}.
Sub-standard housing has more mosquito entry points and most malaria transmissions in sub-Saharan Africa occur inside dwellings while the inhabitants are asleep \citep{tusting2020, tusting2017housing, JATTA2018e498, Tusting2019MappingCI}.
Houses with metal roofs are hotter in the daytime than houses with thatched roofs. This may reduce mosquito survival and inhibit parasite development within the mosquito in metal roof houses. On this basis, the proliferation of modern construction materials in sub-Saharan Africa may have contributed decisively to the reduction of malaria cases \citep{Tusting2019MappingCI}.
Classification of roof characteristics thus holds potential to support malaria surveillance and control programs.
Roof characteristics, such as geometry,  material, and condition can be monitored using remote sensing imagery to advance risk assessment of mosquito-borne diseases and guide mitigation strategies, especially when detailed health and socioeconomic data are scarce. 

Here, we introduce the \ds\ drone-imagery dataset to support the development of machine learning algorithms for automated building \emph{and} roof type mapping in low-income areas prone to malaria risk.
Our dataset is based on high-resolution drone imagery ($\approx4.4$\,cm) of peri-urban and rural settlements in Nacala, Mozambique.
The Mozambican NGO \#MapeandoMeuBairro delineated 17 954 buildings and categorized them according to five roof types, and the authors again carefully verified all annotations.
We define three tasks on the \ds\ dataset,
building detection, multi-class roof type classification, and pixel-level building segmentation.

While these tasks are related, closer inspection reveals a misalignment between their objectives.
Accurate segmentation as measured by the intersection over union (IoU) does not necessarily imply accurate object separation, and vice versa.
For accurate detection and classification, it would be sufficient to only detect the interior of an object as long as the segmented area allows to correctly classify the type. 
If the roofs of two buildings are (almost) touching, then some segmentation may have a high IoU but could make it difficult to separate buildings for counting. This is also a common issue in other applications, e.g., when studying cells in medical images \citep{ronneberger2015unet} or trees from satellite images \citep{brandt:20,mugabowindekwe:22}).

We benchmark three conceptually different  state-of-the-art approaches on our multi-task dataset.
First, we evaluate YOLOv8 \citep{yolov8} developed for object detection, classification, and instance segmentation. 
Second, we build a segmentation model based on \dinov\ \citep{oquab2024dinov}, a state-of-the-art pretrained vision transformer.
Lastly, we evaluate U-Net \citep{ronneberger2015unet} a fully-convolutional encoder-decoder architecture, designed for semantic segmentation.
To address the potential conflicts between pixel-level segmentation and correct object separation as outlined above, we propose a simple approach based on the recent work by \cite{cheng2024scattered}, which we refer to as the Deep Ordinal Watershed (DOW) method. 
We extend both U-Net and DINOv2 to produce an additional output map that predicts the interior of objects.
While the original exterior segmentation map maximizes the IoU, we show that the interior map supports object separation. 

The main contributions of our work are the following:
\begin{enumerate}[nosep,leftmargin=*]
\item We provide the \ds\ dataset containing drone imagery  from peri-urban and rural areas in a sub-Saharan African region. The dataset contains accurate segmentation labels for buildings, categorized into five roof types.
\item Based on the dataset, we define a multi-task machine learning benchmark  for binary and multi-class object detection and semantic segmentation.
We implemented and benchmarked different carefully adopted baseline methods, reflecting three different approaches to address these tasks.
\item We propose a general and simple approach to extend models for semantic segmentation to yield good segmentation \emph{and} object separation results.  
\end{enumerate}
The data and code for reproducing the experiments are made freely available at \url{https://mosquito-risk.github.io/Nacala}.

The next section presents the \ds\ data, provides some background about roof types and risk of vector-borne diseases, and briefly discussed related datasets.
Section \ref{sec:models} describes the deep learning models we evaluated with an emphasis on deep watershed methods. Experimental results are presented in Section \ref{sec:results} before we conclude.

\section{\ds\ Data}\label{sec:data}
\paragraph{Background: Housing conditions and risk of mosquito-borne diseases.}
In sub-Saharan Africa, housing conditions, health outcomes, and socioeconomic status of the residents are interrelated \citep{gram-hansen2019mapping,abraham2019,tusting2020}. As poverty is widespread, diseases are more prevalent, and data are scarce in this region, automatic profiling of housing conditions based on analysis of satellite imagery holds the potential to estimate the socioeconomic status of the inhabitants and assess the risk of disease. This may in turn support targeted public health interventions. 

Mosquitoes are vectors for diseases such as malaria, dengue, Zika, West Nile fever, Chikungunya, and Yellow fever. 
In 2022, more than \num{600000} deaths occurred due to malaria globally and out of the approximately 249 million documented cases, around 233 million occurred within the WHO African Region, accounting for roughly around 94\% of the total documented cases. The economic impact of malaria in Sub-Saharan Africa not only impedes progress towards achieving Sustainable Development Goal~3 (Good Health and Well-being) but also undermines efforts to attain SDG~1 (No Poverty) and SDG~8 (Decent Work and Economic Growth) by compromising economic productivity. Extreme weather conditions caused by climate change will likely exacerbate problems with mosquito-borne diseases in sub-Saharan Africa, as floods are expected to increase in frequency and have been linked to outbreaks of malaria in Africa \citep{10665-268220}.

Low-quality housing increases the risk of transmission of diseases by mosquitoes, as sub-standard houses have more mosquito entry points and thereby increase human exposure to infection in the home~\citep{tusting2015,nomcebo_2017}. Mosquito survival is lower in metal-roof houses compared to thatched-roof houses due to higher daytime temperatures \citep{tusting2015}. Most malaria transmissions in sub-Saharan Africa occur indoors at night, and poor climatic performance of housing has been linked to increased malaria risk \citep{JATTA2018e498}. This is because elevated indoor temperatures can cause discomfort for inhabitants, which may result in decreased use of mosquito nets during the night. Roof materials, geometry, and conditions are critical for indoor climate, as roofs comprise the primary surface exposed to the sun. Automatic classification of roof characteristics thus holds potential for informing risk assessment of malaria and support targeted interventions.

\paragraph{The \ds\ dataset.}
We gathered drone imagery of the Nacala region in Mozambique.
The burden of malaria in Mozambique is approximately 10-fold the world average (number of documented cases compared to the total population, \citealp{VENKATESAN2024e214}).
The data covers three informal settlements of Nacala, a city of \num{350000} inhabitants on the northern coast of Mozambique. 
Aerial imagery was collected using a DJI Phantom 4 Pro drone and processed using AgiSoft Metashape software. 
All data was recorded between October and December 2021, under a development project led by \#MapeandoMeuBairro and supported by Nacala Municipal Council. 
The image resolution is $\approx4.4$\,cm, and we made all raw imagery available in OpenAerialMap~\citep{openaerialmap}.
The total number of buildings in the study areas is \num{17954}. 
We distinguished five major types of roof materials in Nacala, namely metal sheet, thatch, asbestos, concrete, and no-roof, and their counts are \num{9776}, \num{6428}, \num{566}, \num{174}, and \num{1010}, respectively. 
The region is mostly dominated by metal sheets and thatch roofs.

\begin{figure}[ht]
\centering
\includegraphics[width=0.99\linewidth]{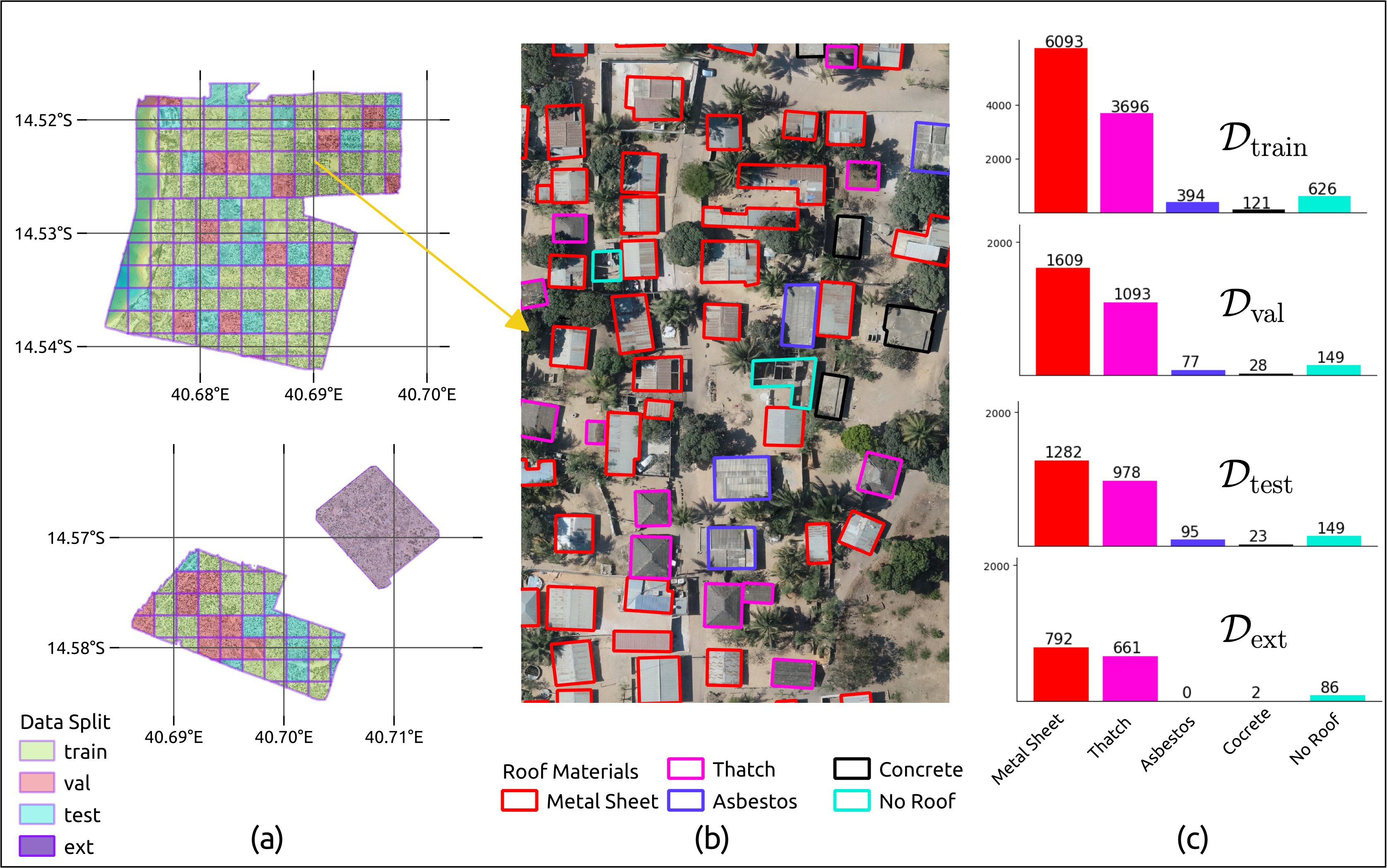}
\caption{(a) Visualisation of the training, validation and test sets with reference to longitude and latitude; (b) Drone imagery with labels; (c) Instance counts for each class in all sets.}
\label{data} 
\end{figure}

From the three informal settlements, see Figure~\ref{data}, the first two areas were split into training \dtrain, validation \dval, and test \dtestone\ using stratified sampling. 
We created a square grid of 225 meters and counted the roof types in these cells. Then we partitioned the cells into three sets based on the class counts to achieve a similar class distribution in each set, where we prioritized the distribution of minority classes (i.e., concrete and asbestos). 
We defined that a building only belongs to a specific grid cell if its centroid falls into the cell. 
If a building area falls into two grid cells and those two cells belong to two different sets (e.g., training and test set), we choose to have data pixels in the set where the centroid of the building is placed. The remaining part of the building in the other set was masked to avoid data leaking between sets. 

Although objects in training, validation, and test sets are from different cells, they stem from the same two areas. 
To evaluate the generalization to a new area without adjacent training data, we hold out the third settlement as a second test set referred to as \dtesttwo.

\paragraph{Related datasets.}
The project ``Mapping Informal Settlements in Developing Countries using Machine Learning with Noisy Annotations and Multi-resolution Multi-spectral Data'' \citep{helber2018generating,gram-hansen2019mapping} is most closely related to our work. 
They used freely available 10m/pixel resolution imagery from the Sentinel-2 satellite and obtained labels for three roof types (metal, shingles, thatch)
from geo-located survey data provided by Afrobarometer\footnote{\url{www.afrobarometer.org}}. These labels are very noisy in space and time. The labels are often not aligned with buildings because the geo-located coordinates were distorted for privacy reasons. Furthermore, the survey questions and satellite image observations may not be aligned in time. 
While the low spatial resolution of the Sentinel-2 imagery might allow to cover large geographic regions, it makes roof type classification challenging \citep{helber2018generating}. 

There are many datasets that contain remote sensing imagery with building labels, which, however, typically do not distinguish roof types.
In particular, \emph{Open Buildings} is a freely available continental-scale building dataset covering the whole of Africa
\citep{sirko2021continental}. In comparison,  \ds\ is much more focused, providing significantly higher resolution images, more accurate delineations, and in particular roof type classifications.

\cite{alidoost18} distinguish between roof types in aerial images. However, they map a rather high-income town in Germany, where they distinguish between 
three roof shapes common in that region (flat, gable, and hip). Another dataset for classifying roof geometry is provided by \cite{persello2023}, who distinguish 12 fine-grained details of roof geometry.

\section{Benchmarked Methods}\label{sec:models}
This section presents the approaches we benchmarked on the \ds\ data set.
The goal is to accurately segment the buildings (as assessed by metrics based on the IoU), separate individual buildings, and classify the roof materials.
As baselines, we considered U-Net~\citep{ronneberger2015unet}, YOLOv8~\citep{yolov8}, and a model performing segmentation based on \dinov~\citep{oquab2024dinov}. 
Furthermore, we extend the U-Net and the \dinov{} based systems with the deep ordinal watershed method recently proposed by \cite{cheng2024scattered}.
These approaches are compared in two settings. 
In the \emph{two-stage} setting, we first solved the building segmentation and separation tasks and afterwards  classified  the roof material for each detected building. In the \emph{end-to-end} setting, segmentation and classification were done in parallel.

\subsection{Baseline Models}
\begin{figure}[!tbp]
    \centering
    \resizebox{.99\linewidth}{!}{\input{figures/Figure_d_arXiv}}
    \caption{Baseline (top) and DOW (bottom) variants of our systems using either ResNet35 (in the case of the U-Net architectures) or DINOv2 as encoders.     
    When using DOW, The watershed algorithm takes two segmentation masks as input, the predicted objects (level 1) and their interiors (level 2).
    In the two-stage approach, the classifier shown in Figure~\ref{dinovc} is using the binary building segmentation (left). In the end-to-end setting, the roof material is predicted directly with a multi-class segmentation approach (right). 
    }
    \label{fig:architectures} 
\end{figure}
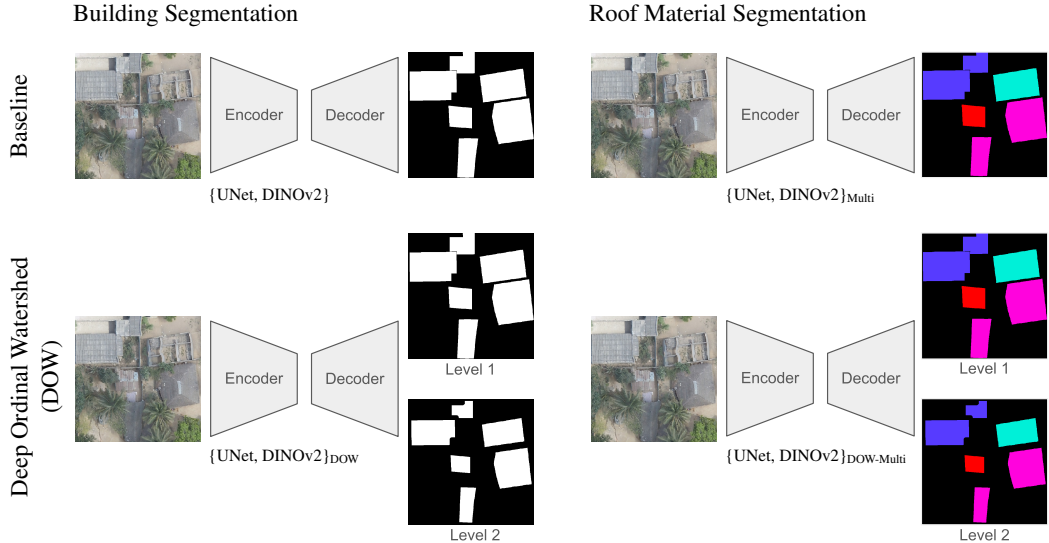

\paragraph{U-Net.}
The U-Net is arguably the most common architecture 
for semantic segmentation~\citep{ronneberger2015unet}.
We utilized a  ResNet34~\citep{resnet} encoder pretrained on ImageNet and a decoder similar to the original U-Net, except that we used nearest-neighbor upsampling instead of transposed convolutions \citep{nn-upsampling}, see Figure~\ref{unet} in the Appendix.

To identify individual instances in the semantic segmentation output map, the connected components in the map were determined \citep{brandt:20}.
To better separate individual buildings, we used a pixel-wise weight map during training that puts more emphasis on the space between buildings as already suggested by \citeauthor{ronneberger2015unet}
(see Appendix~\ref{sec:unet_weights} for details) and commonly used in remote sensing \citep[e.g.][]{brandt:20}.
However, this is not sufficient to separate buildings that are very close to each other or touch each other.
Thus, we modified the target segmentation masks during training: Some border pixels were relabeled as background to ensure that there is a minimum gap of $\ngap=7$ pixels between roofs. This modification of the target masks was only applied during training, before computing the weight map but not when calculating any performance metrics.

\paragraph{YOLOv8.}
We trained YOLOv8 
\citep{yolov8}, which is among the state-of-the-art methods for instance segmentation. 
We fine-tuned a model pretrained on the COCO dataset.
While the original YOLO architecture was designed for object detection, YOLOv8 allows for instance segmentation by integrating concepts from YOLACT~\citep{bolya2019yolact}.

\paragraph{DINOv2.}
We benchmarked an approach based on DINOv2~\citep{oquab2024dinov}, a state-of-the-art pretrained vision transformer. It uses the DINOv2 \emph{Base} model as an encoder, which is extended by a convolutional decoder.
The DINOv2 output, a patch embedding with the shape of $\mathbb{R}^{1024 \times 768}$, is reshaped into feature maps of size $\mathbb{R}^{32 \times 32 \times 768}$. 
Then convolutional and linear upsampling layers are used on top of these feature maps as a decoder (see Appendix~\ref{dinovs-arc}). 
We used the  same loss function, weighting function, training label adjustment, and training strategy as for U-Net. We froze the encoder weights and only the convolutional decoder was trained.

\subsection{Deep Ordinal Watershed}
U-Nets and the \dinov{} based method described above try to classify each pixel as accurately as possible. 
However, for proper separation of objects it is sufficient -- and typically preferable -- if only the interior of an object is segmented.
If the border of a building can be classified as background, even touching buildings can be separated.
This reasoning leads to the deep ordinal watershed  (DOW) model introduced by \cite{cheng2024scattered}.

In the watershed approach, each pixel is assigned a height and the image is viewed as a topological map \citep{soille1990automated}.
A DOW architecture does not only predict a single segmentation mask but $\nl$ feature maps  
for $\nl+1$ discrete height levels, $\{0,1,\dots,\nl\}$, where $0$ corresponds to the highest and $\nl$ to the lowest elevation. Background pixels are assumed to have  level 0. The Euclidean distance transformation is computed for each object, and
the distances are discretized into the remaining $\nl$ height levels.
Target feature map $m\in\{1,\dots,\nl\}$ marks all pixel with a distance level of 
$m$ or higher. 
That is, the objects in the target feature maps get smaller with increasing $m$
(if $\nl=1$ we recover the standard U-Net).  
Learning the discrete height levels of  pixels this way solves  an ordinal regression task \citep{frank2001simple,cheng2008neural}.
Given the pixel heights, the watershed algorithm can be applied as a post-processing step for instance segmentation \citep{soille1990automated}. 
Local minima in the elevation map define basins, each of which defines a distinct object.
Adopting a flooding metaphor, the watershed algorithm now floods the basins until basins attributed to different starting points meet on watershed lines. Pixels attributed to the same basin belong to the same object.

\cite{cheng2024scattered} employ a DOW U-Net for individual tree segmentation, however, without a comparison with a standard U-Net or exploring different numbers of levels. For our task, we hypothesize 
that a  minimal number of $\nl=2$ different non-background heights is sufficient.
In this setting, the system outputs two masks representing the full object and its interior, respectively. Let $\np$ denote the difference in distance between two levels.
The smallest building in our data set has  size \SI{1.463}{\metre^2}. Thus, for the given image resolution, the number of pixels per side is approximately $\sqrt{1.46}\big/0.044$. This suggests to define the levels such that  $\np < 13$, and we picked $\np=10$.

We empirically evaluated DOW variants of both our U-Net and \dinov{} based systems, 
see  Figure~\ref{fig:architectures}.
We describe the U-Net extension in more detail in Appendix~\ref{uneth-arc}, the \dinov{} based systems were modified analogously. The  DOW U-Net network architecture \uneth{} used in our study is illustrated in Figure~\ref{uneth} in the appendix. For a comparison with a DOW U-Net with $\nl=6$ we refer to Appendix~\ref{uneth-arc} and Appendix~\ref{app:results}.

Although the approaches are related, we would like to stress the DOW method is  conceptually different from \emph{deep level sets}, where deep neural networks learn a (continuous) level set function, the zero-set of which defines object boundaries
\citep{hu2017deep,Hatamizadeh20}, as well as from predicting interior and border of an object as, for instance, done by \cite{girard2021polygonal}.

\subsection{Two-stage vs.~End-to-end}\label{sec:twostage}
\begin{figure}[ht]
    \centering
    \includegraphics[width=0.99\linewidth]{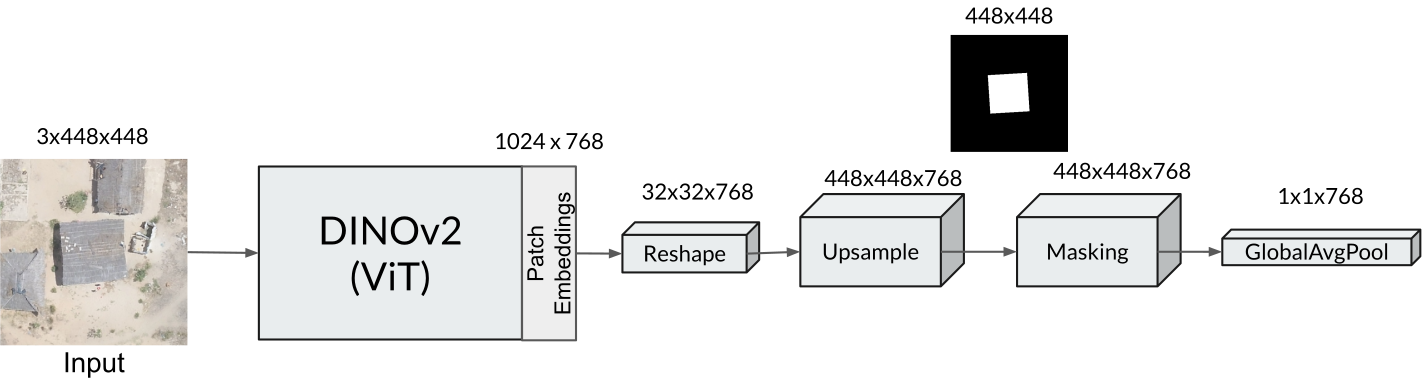}
    \caption{The architecture of the DINOv2 based roof material classifier used in the two-stage setting. A classifier (e.g., logistic regression) is applied to the resulting feature vector.}
    \label{dinovc} 
\end{figure}

All the neural network architectures described above can directly classify the roof types of detected buildings by predicting multi-class segmentation masks. 
However, encouraged by good classification results using  DINOv2 features, we also studied an alternative two-stage approach: First we segmented and separated the buildings using the  algorithms described above ignoring the roof material information. 
That is, we reduced the multi-class problem to a binary task.
After that, we predicted the roof material of each detected building. 
We used DINOv2 to processes a $448 \times 448$ patch centered around each building, 
see Figure~\ref{dinovc}. 
The output of DINOv2, a patch embedding with the shape of $\mathbb{R}^{1024 \times 768}$ was reshaped into feature maps of $\mathbb{R}^{32 \times 32 \times 768}$. 
These feature maps were then upsampled to the input patch size, masked with a target binary building mask, and average pooling was applied to obtain the final feature vector for the building. 
Standard machine learning classifiers were applied to this embedding to predict the roof material, where linear probing gave the best results (see Appendix \ref{app:classifier} for a comparison of different classifiers).

The two-stage methods are referred to as U-Net, DINOv2, \uneth, and \dinovh{}, and the corresponding end-to-end methods are denoted by \unetm, \dinovsh, \unethm, and \dinovsh,
see Figure~\ref{fig:architectures} for an overview.

\section{Experiments and Results}\label{sec:results}

\subsection{Experimental Setup}
All models, except for YOLOv8 where we followed its original training protocol, were trained using cross-entropy loss with pixel-wise weighting. 
We employed the AdamW optimizer~\citep{adamw} with an initial learning rate of $0.0003$.  
All models were trained for 300 epochs, utilizing a learning rate scheduler that decreased the learning rate by a factor of $10$ every $50$ epochs. 
The final weight configuration and hyperparameters for each model were selected based on the highest IoU score achieved on the validation dataset.
The hyperparameters of the U-Net were chosen by observing results on the validation data set in an iterative process. 
The high training speed of \yolob{} allowed for more systematic model selection: We applied the genetic algorithm that comes as part of the \yolob{} framework for hyperparameter optimization \citep{yolov8}.
The input patch sizes for the U-Net variants, YOLOv8, and DINOv2 models were 512, 640, and 448, respectively. 

\subsection{Evaluation Metrics}
The semantic segmentation performance was evaluated by the  IoU.
We considered both the IoU of the binary building segmentation and the mean IoU for class-specific roof segmentation.  
The roof materials concrete and asbestos are very rare. While \dtrain, \dval, and \dtestone{} are stratified samples containing all classes, the spatially distinct data \dtesttwo{} does not contain any example of the two roof types, see Figure~\ref{data}.
To allow for a better comparison between the two test sets and to see the effect of the rare classes on the macro-averaged mean IoU, we provide the mean IoU of the three main classes (\iout{}) alongside with the mean IoU of all five classes (\iouf{}).

Instance segmentation was assessed using the  \apf{} score, that is, the
average precision evaluated at an IoU threshold of $0.5$ \citep{pascal-voc-2007,lin2014microsoft}.  
We evaluated the AP for both the predictions of building instances and the predictions of multi-class roof type instances (i.e., in the latter case an object is only detected if the roof material is correctly identified). 
Similar to IoU, \apft{} and \apff{} denote the mean \apf{} over three and five classes.
To estimate the average precision, a confidence score is required for each building segment. 
The confidence score of binary and multi-class segmentation models was obtained by 
interpreting the neural networks' outputs as probability distributions over classes and calculating the mean probability of belonging to the predicted class over all pixel within a predicted segment.
The exception was YOLOv8, which provides its own confidence score.
When a classifier using \dinov{} features was used on top of binary segmentation models, the confidence score was derived from the canonical probability score of the classifier. 
Additional metrics,  \apfn{} and \tps{}, are shown in Appendix~\ref{app:results}.
Information on the computer resources  is provided in Appendix~\ref{app:compute}.

\subsection{Results and Discussion}

\begin{table}[!tbp]
  \centering
  \caption{
  {Benchmarking results on the \ds\ dataset.}
  The table reports averages over five trials $\pm$ standard deviations. 
  The upper five models were trained in the two-stage setting. The lower half of the models was trained in the end-to-end setting, where multi-class classification is performed together with the segmentation as indicated by the subscript \emph{Multi}. Models that used the DOW extension are indicated by the subscript \emph{DOW}. 
  IoU and \apf{}  were computed on the binary output, where  the predictions of multi-class models were binarized. mIoU and \mapf{} are macro averages, the superscipts indicate whether the averaging was done over all five classes or over  the three frequent roof types. Results  for individual roof types can be found in Appendix~\ref{app:results}.}
  \resizebox{\textwidth}{!}{%
  \begin{tabular}{@{}lrrrrrrrrrr@{}}
    \toprule
    & \multicolumn{6}{c}{\dtestone}  & \multicolumn{4}{c}{\dtesttwo} \\
    \cmidrule(lr){2-7} \cmidrule(lr){8-11}
    & \multicolumn{3}{c}{pixel level}  & \multicolumn{3}{c}{object level} & \multicolumn{2}{c}{pixel level} & \multicolumn{2}{c}{object level} \\
    \cmidrule(lr){2-4} \cmidrule(lr){5-7} \cmidrule(lr){8-9} \cmidrule(lr){10-11}

    Model Name  & IoU & \iout & \iouf & \apf & \apft & \apff & IoU & \iout & \apf & \apft \\
    \midrule
    \yolob & \ccell{0.866}{0.012} & \ccell{0.713}{0.019} & \ccell{0.568}{0.015} & \bcell{0.941}{0.003} & \ccell{0.815}{0.011} & \ccell{0.698}{0.018} & \ccell{0.896}{0.002} & \ccell{0.761}{0.006} & \bcell{0.963}{0.005} & \ccell{0.846}{0.008} \\
    \addlinespace[2pt]

    DINOv2 & \ccell{0.833}{0.002} & \ccell{0.755}{0.004} & \ccell{0.562}{0.003} & \ccell{0.882}{0.004} & \ccell{0.789}{0.006} & \ccell{0.683}{0.008} & \ccell{0.905}{0.000} & \ccell{0.747}{0.011} & \ccell{0.919}{0.005} & \ccell{0.806}{0.008} \\
    \addlinespace[2pt]

    \dinovh & \ccell{0.884}{0.001} & \ccell{0.763}{0.002} & \ccell{0.565}{0.004} & \ccell{0.930}{0.005} & \ccell{0.836}{0.002} & \ccell{0.725}{0.004} & \ccell{0.905}{0.001} & \ccell{0.852}{0.007} & \ccell{0.956}{0.001} & \ccell{0.852}{0.007} \\
    \addlinespace[2pt]
    
    U-Net & \bcell{0.895}{0.003} & \ccell{0.757}{0.024} & \ccell{0.570}{0.016} & \ccell{0.910}{0.005} & \ccell{0.810}{0.008} & \ccell{0.688}{0.014} & \ccell{0.909}{0.001} & \ccell{0.748}{0.007} & \ccell{0.929}{0.000} & \ccell{0.787}{0.011} \\
    \addlinespace[2pt]
    
    \uneth & \bcell{0.895}{0.002} & \ccell{0.775}{0.013} & \ccell{0.577}{0.009} & \ccell{0.935}{0.001} & \ccell{0.836}{0.005} & \bcell{0.730}{0.011} & \bcell{0.911}{0.002} & \ccell{0.764}{0.006} & \ccell{0.947}{0.004} & \ccell{0.812}{0.008}\\
    \addlinespace[2pt]
    
    \midrule
    
    \yolom & \ccell{0.824}{0.023} & \ccell{0.708}{0.010} & \ccell{0.550}{0.017} & \ccell{0.910}{0.005} & \ccell{0.816}{0.009} & \ccell{0.597}{0.007} & \ccell{0.885}{0.002} & \ccell{0.785}{0.006} & \ccell{0.948}{0.003} & \ccell{0.849}{0.015} \\
    \addlinespace[2pt]
    
    \dinovs & \ccell{0.880}{0.002} & \ccell{0.774}{0.004} & \ccell{0.699}{0.012} & \ccell{0.899}{0.003} & \ccell{0.820}{0.010} & \ccell{0.689}{0.025} & \ccell{0.899}{0.002} & \ccell{0.818}{0.005} & \ccell{0.946}{0.001} & \bcell{0.880}{0.011} \\

    \dinovsh & \ccell{0.885}{0.001} & \bcell{0.786}{0.006} & \bcell{0.734}{0.006} & \ccell{0.918}{0.003} & \ccell{0.824}{0.011} & \ccell{0.702}{0.013} & \ccell{0.902}{0.001} & \bcell{0.819}{0.006} & \ccell{0.950}{0.005} & \ccell{0.875}{0.010} \\
    
    \unetm & \ccell{0.879}{0.012} & \ccell{0.783}{0.010} & \ccell{0.634}{0.024} & \ccell{0.924}{0.004} & \bcell{0.850}{0.011} & \ccell{0.716}{0.018} & \ccell{0.903}{0.002} & \ccell{0.805}{0.020} & \ccell{0.943}{0.010} & \ccell{0.844}{0.039} \\
    \addlinespace[2pt]
    
    \unethm & \ccell{0.892}{0.001} & \ccell{0.777}{0.012} & \ccell{0.672}{0.042} & \ccell{0.928}{0.002} & \ccell{0.829}{0.011} & \ccell{0.671}{0.022} & \ccell{0.904}{0.002} & \ccell{0.794}{0.014} & \ccell{0.942}{0.005} & \ccell{0.812}{0.021} \\
    \addlinespace[2pt] 
    \bottomrule
  \end{tabular}}
  \label{new_table1}
\end{table}

Our experimental results on \dtestone{} and \dtesttwo{} are presented in Table~\ref{new_table1}, additional details can be found in Appendix~\ref{app:results}. All metrics on the test sets were computed on raw images instead of patches to avoid artifacts when splitting images.
We report averages over five trials on the corresponding standard deviations.
The methods reached \apf{} and IoU values on the spatially separated test set of up to $0.963$ and $0.880$, respectively. Thus
the tasks can be solved with an accuracies high enough for subsequent analysis while  still leaving room for improvement. Detecting thatch roofs is particularly relevant, as they are associated with an increased malaria risk \citep{Tusting2019MappingCI}, and these roofs can be identified particularly well, see Table~\ref{object-level} in the Appendix.

\begin{figure}[!htbp]
    \centering
    \includegraphics[width=0.95\linewidth]{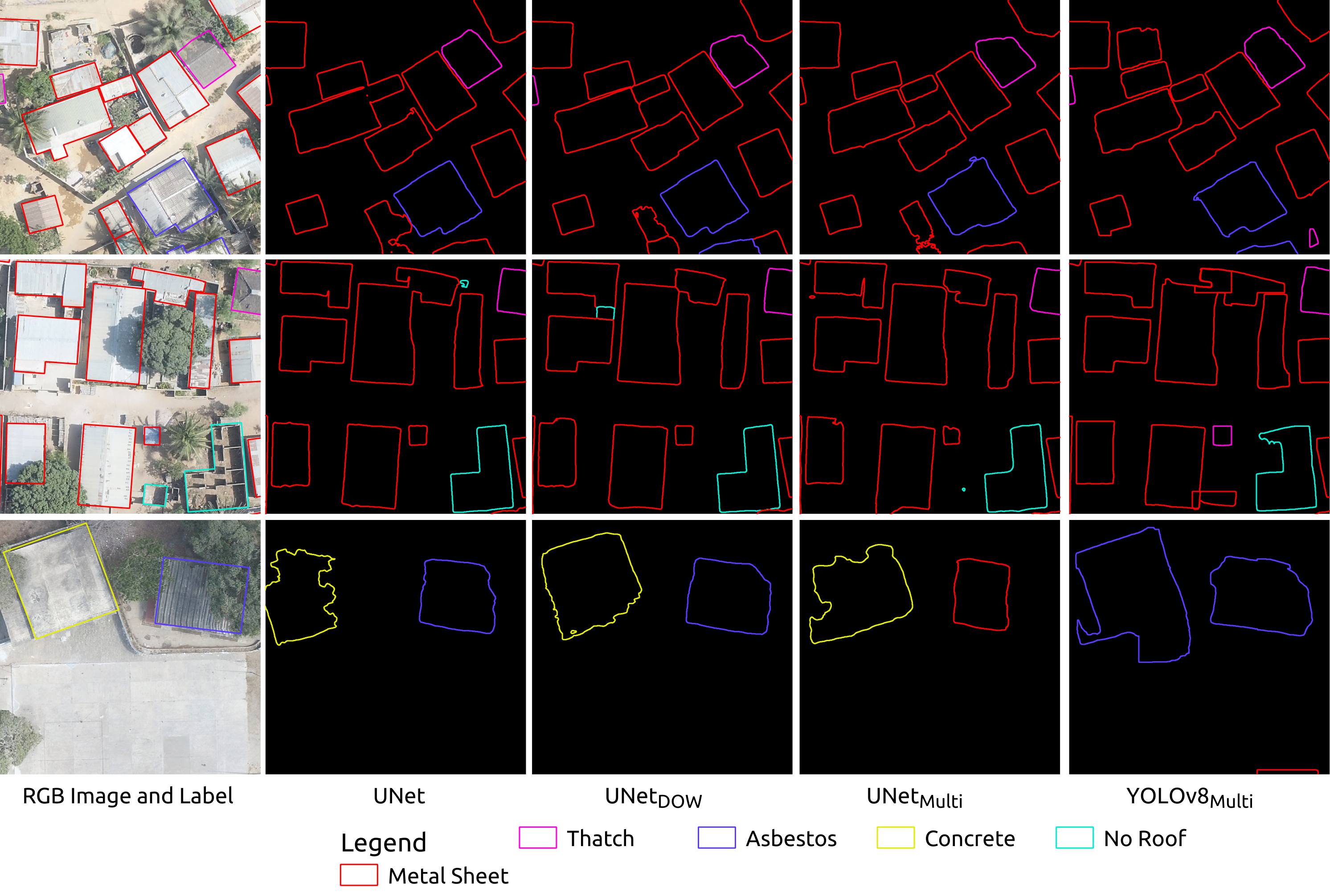}
    \caption{Exemplary predictions on \dtestone{} by different models. The predictions are polygonized and colored by class. The roof types with few training examples, asbestos and concrete, are particularly difficult, see bottom row.} \label{results} 
\end{figure}

When comparing the different approaches, we find that there is no method that was better than the others across all metrics.
The U-Nets and \yolob{} did well on their home grounds: \yolob{} gave good object detection results (e.g., the best \apf{} scores), while the U-Nets performed well for semantic segmentation as measured by IoU. \dinov{} combined with a simple decoder was also competitive.
Exemplary results are shown in Figure~\ref{results}. 
As could be expected, classifying the minority roof types asbestos and especially concrete (which resembles concreted background areas)
was most difficult, in particular for end-to-end \yolob{}, see Table~\ref{object-level}. 
\yolob{} had the tendency to produce
artefacts when applied to the larger images.
This is one of the reasons for its lower IoU score.

In general, the DOW extension improved
both  U-Nets and  \dinov{} based architectures. 
Comparing \dinov{} with \dinovh{} and U-Net with \uneth, the DOW variants were better in all ten performance indices (except for IoU on \dtestone{} where U-Net and \uneth\ gave the same result). 
Comparing \dinovs\ with \dinovsh, the latter was better in all indicators except \apft\ on \dtesttwo. Only for \unethm\ the results were mixed, using DOW gave lower values for five indices and higher values for the other half. 
Overall, the DOW extension had a statistically significant positive effect on the object separation as intended. If we pool all 20 DOW trials and compare with the corresponding trials predicting a single mask, then the AP50 improved significantly (two-sided Wilcoxon rank sum test, $p<0.001$) while the difference in IoU was not significant ($p>0.05$).

\paragraph{Limitations.}
The \ds\ dataset is not a large-scale data set by current standards and it is restricted to a single region. However, considering the proliferation of low-cost drone technologies, high-resolution geospatial surveying is becoming increasingly affordable and common in sub-Saharan Africa. Accordingly, similar but unlabelled data will likely become available in the coming years at large scale, which makes it important to develop methods to make good use of these data now.
The \ds\ dataset covering informal settlements is a good example for the target areas of our risk disease monitoring and prevention research. In this context, Mozambique is particularly relevant because the country suffers from a high malaria incidence rate \citep{VENKATESAN2024e214}.
The second test set allows for testing generalization in an area geographically separated from the main training/test/validation data.
In general, we would argue that there is a need for medium size 
benchmark data sets such as the \ds\ data to support equity in machine learning research, as we need benchmarks that can be utilized by researchers with limited compute resources. 

\section{Conclusions}\label{sec:conclusions}
The \ds{} dataset contains high-resolution drone imagery from informal settlements in Mozambique, where buildings  and their  roof material were carefully annotated.
We curated the dataset as part of an intercontinental and interdisciplinary research project on risk assessment of mosquito-borne diseases, especially malaria, with the goal to predict risk maps and to develop and support measures for risk reduction.
From a methodological perspective, the dataset defines a multi-task problem.
We are interested in accurate semantic segmentation to determine the roof areas and also in identifying the individual buildings and classifying their roof types. Thus, the dataset adds to the landscape of computer vision benchmarks by providing a relevant resource for the development and evaluation of frameworks that strive at solving semantic segmentation as well as object detection and classification simultaneously with a high accuracy.
For example, working on the \ds{} data has led us to the proposed deep ordinal watershed (DOW) approach, a reduced variant of the method described by \cite{cheng2024scattered}. This variant method first segments objects along with their interiors into two elevation levels and then performs a watershed segmentation to separate objects. The DOW idea is applicable beyond the \ds\ data, on which it improved both the standard U-Net architectures as well as a system based on DINOv2 features for segmentation. 
Implementations of all algorithms are made publicly available together with the data (\url{https://mosquito-risk.github.io/Nacala}). 
With the \ds\ dataset, we invite the machine learning community to develop new approaches for interpreting high-resolution drone images that can ultimately support risk assessments of vector-borne diseases.

\acksection
We acknowledge support through the project 
\emph{Risk-assessment of Vector-borne Diseases Based on Deep Learning and Remote Sensing} (grant number NNF21OC0069116) by the Novo Nordisk Foundation.

{
\small

\bibliographystyle{abbrvnat}
\bibliography{main}
}

\clearpage
\appendix

\renewcommand\thefigure{\thesection.\arabic{figure}} 
\renewcommand\thetable{\thesection.\arabic{table}} 
\renewcommand\theequation{\thesection.\arabic{equation}}

\appendix

\section{Details on Models and Training Procedure}

\subsection{U-Net}\label{unet-arc} \label{sec:unet_weights}
\begin{figure}[!htbp]
    \centering
    \includegraphics[width=0.7\linewidth]{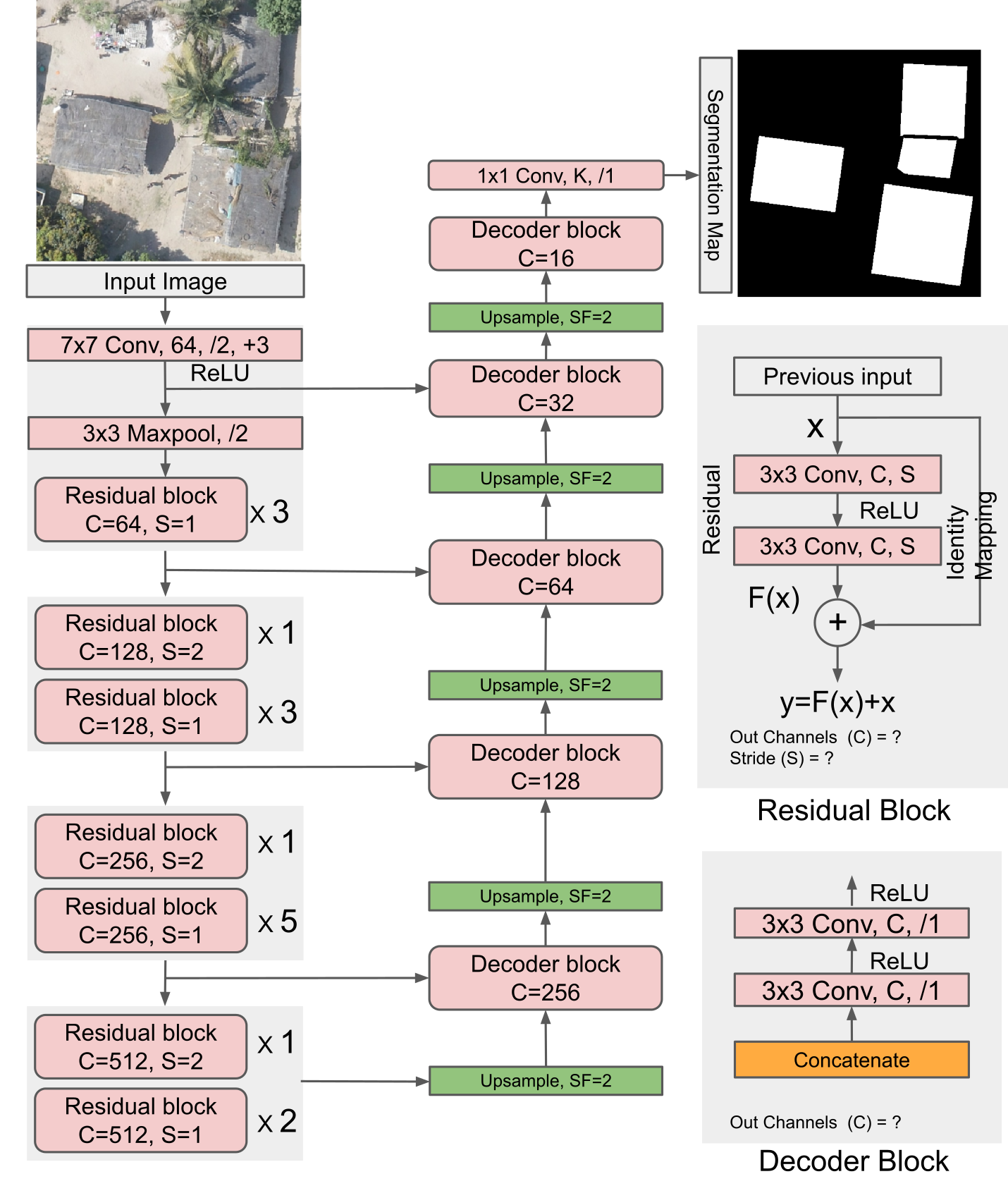}
    \caption{Basic U-Net architecture}
    \label{unet} 
\end{figure}
The basic U-Net architecture we used is shown in Figure~\ref{unet}.

During training, the loss of each background pixel $x$ is multiplicatively weighted 
by $w(x)$ defined as
\begin{equation}
    w(x) =  w_0 \cdot \exp \left( -\frac{(d_1(x) + d_2(x))^2}{2\sigma^2} \right ) \label{unet_weights}
\end{equation}
following \cite{ronneberger2015unet}.
Here,  $d_1 (x)$ denotes the distance to the border of the nearest segment, and $d_2 (x)$ is the distance to the border of the second nearest segment.  We set $w_0 = 10$ and $\sigma = 5$ according to~\cite{ronneberger2015unet}.

During training, we modified the target masks to ensure that $d_1(x) + d_2(x)\ge\ngap=7$ for each background pixel $x$ before we computed the weights $w(x)$.

\subsection{Deep Ordinal Watershed U-Nets}\label{uneth-arc}

\begin{figure}[!htbp]
    \centering
    \includegraphics[width=0.8\linewidth]{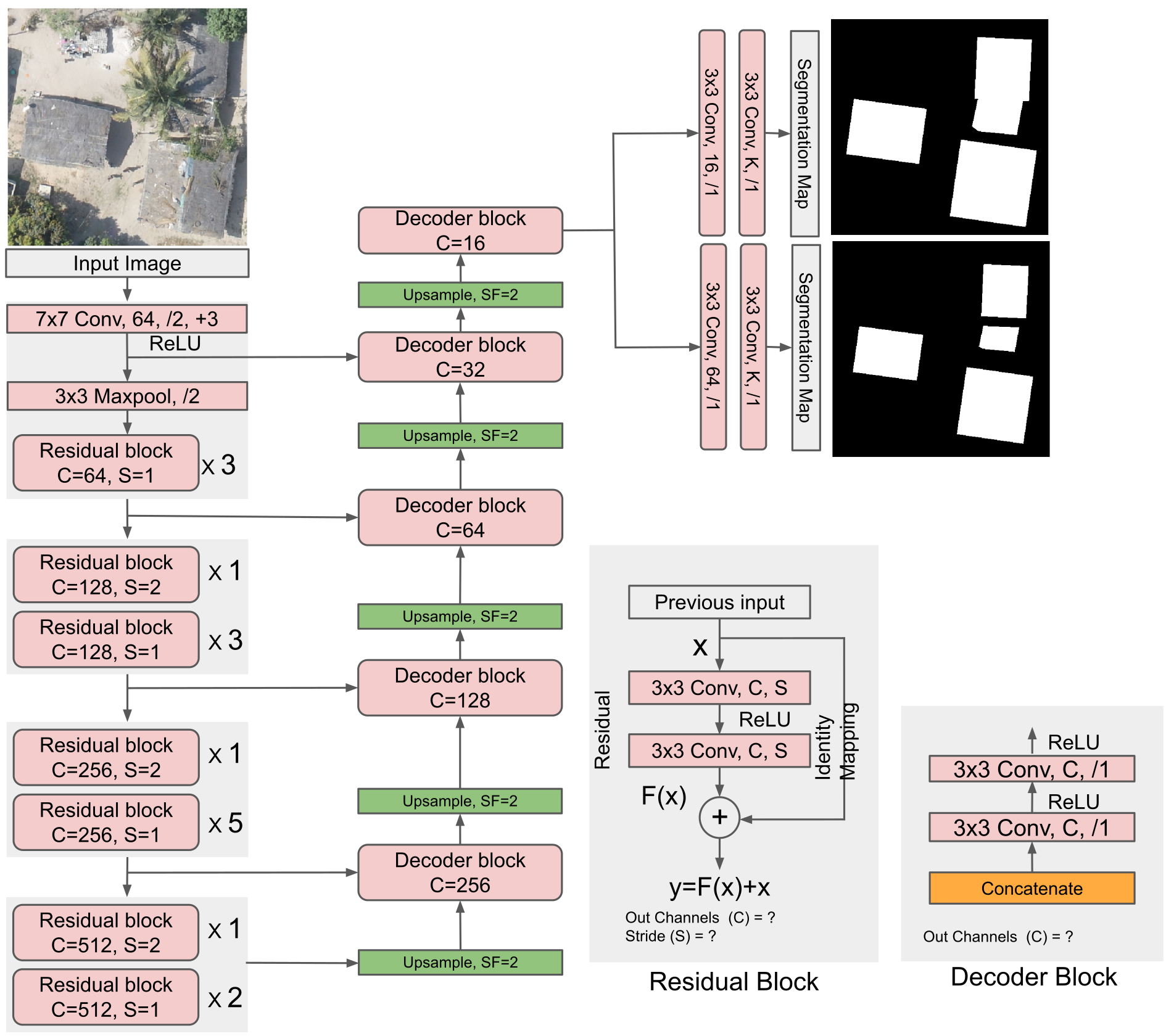}
    \caption{\uneth\ architecture producing two output maps, segmenting objects and their interiors, respectively. The architecture differs from the baseline U-Net only in the output heads.}
    \label{uneth} 
\end{figure}

We considered a stripped down version of the DOW U-Net proposed by \cite{cheng2024scattered} and set the number of elevation levels to $\nl=2$.
The architecture of the resulting DOW network is depicted in Figure~\ref{uneth}, which extends the basic U-Net architecture shown in Figure~\ref{unet}.
In contrast to the original U-Net, the DOW model has two heads. One is predicting an object's area, while the other predicts its interior. The interior is defined by removing pixels within a 10-pixel distance from the border of the building segment.
Each head comprises a convolutional layer, batch normalization, ReLU activation, and finally a pointwise convolutional layer with outputs equal to the number of classes.
While the first head had filters of size $3 \times 3$ in its first convolutional layer, the second head for the interior used  64  filters. The class label of an object was derived from the second head. If no interior was predicted, which can happen in the case of small objects, the output from the first head defined the class.

We compared this DOW variant, referred to as \uneth{},
to the original DOW with several elevation levels, in which the  levels are added to the standard U-Net  architecture (Figure~\ref{unet}) simply by increasing the number of output masks.  
We considered $\nl=6$ discrete height levels and accordingly refer to the model as \unete. 
The pixel margin $\np$ for each height level was determined experimentally by testing  $\np\in\{ 1, 3, 5, 7, 9, 11, 13,  15\}$ on validation data, leading to $\np=5$ for \unete. 
An experimental comparison of \uneth{} and  \unete{} can be found in the extended results in Section~\ref{app:results} in the appendix.

\subsection{Segmentation and Classification Using DINOv2}\label{dinovs-arc}
\begin{figure}[!htbp]
    \centering
    \includegraphics[width=0.8\linewidth]{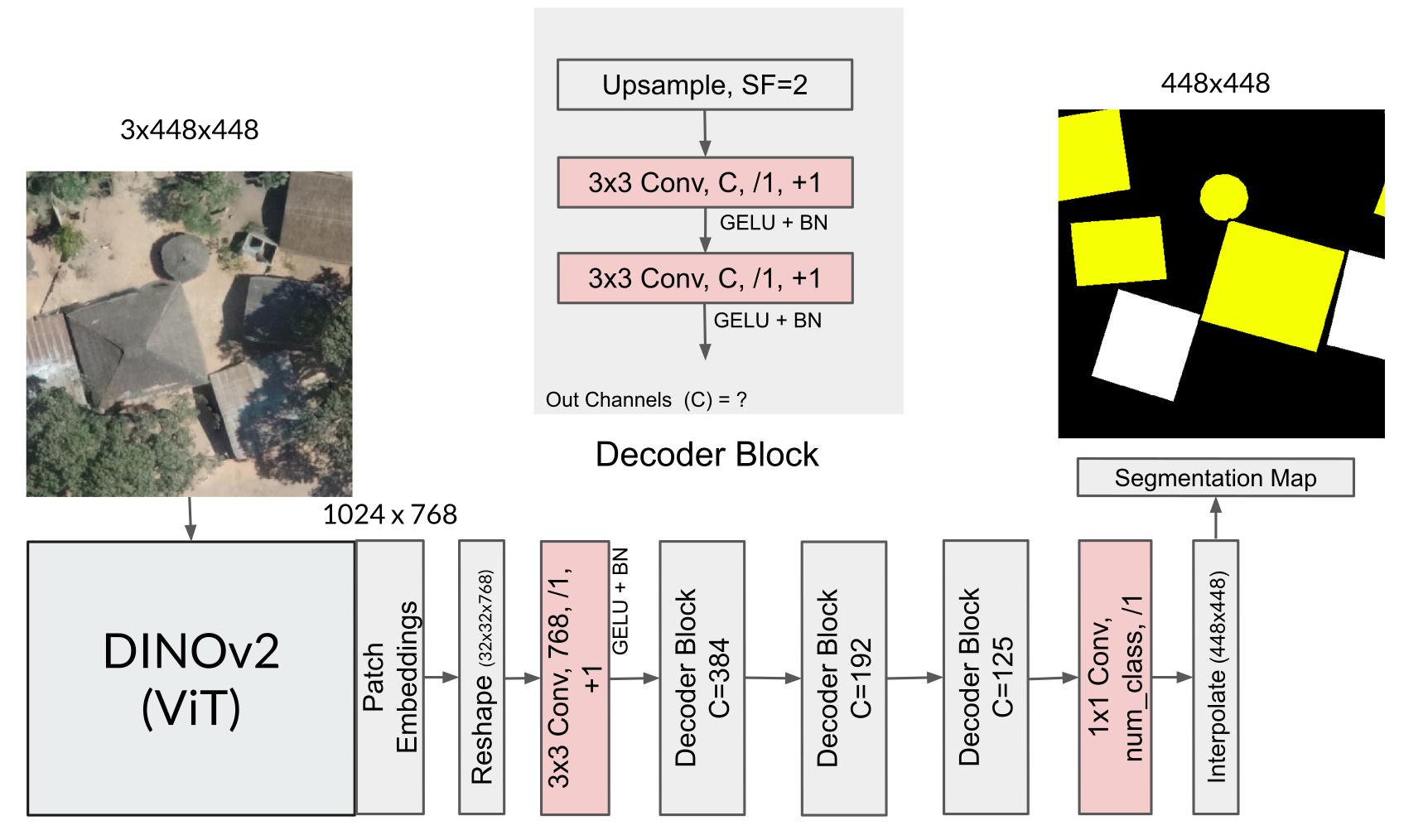}
    \caption{DINOv2 architecture}
    \label{dinovs} 
\end{figure}

The segmentation architecture based on DINOv2 is illustrated in Figure~\ref{dinovs}. 
We refer to it simply as DINOv2. From this architecture, we derived \dinovh{} in the same way as we extended U-Net to  \uneth{} . 

\subsection{Compute resources}\label{app:compute}
All experiments were conducted on AMD MI250X GPUs 
with 64\,GB VRAM provided by LUMI\footnote{\url{https://lumi-supercomputer.eu}}. A total of 8550 GPU hours were used for the project, including preliminary experiments not included in the paper. The computation time for training semantic segmentation model was approximately 20 hours for 300 epochs when the entire data were loaded to GPU memory.

\section{Additional Results}\label{app:results}

\subsection{Detailed Results for Different Roof Materials} \label{sec:dtestone}
Additional results on \dtestone{} are presented in  Table~\ref{pixel-level} and Table~\ref{object-level}.  The tables report the IoU scores for the individual roof material classes. They also show the true positive rates \tps{} in addition to the \apf{} the \apfn{}.
The \apfn{} is defined as the mean AP over IoU thresholds from $50\,\%$ to $95\,\%$ with an interval of $5\,\%$.
The mean of \apfn{} over all classes is \mapfn{}.
\tps{} are the number of segments that overlap with ground truth segments with a minimum IoU of $0.5$, we used this metric to assess the counting of buildings.

Beyond the performance metrics already discussed, we have included the results for  \unete{} as described in Section~\ref{uneth-arc} in the appendix, showing that the two DOW architectures perform on par.

The corresponding results on \dtesttwo{} are given in Table~\ref{pixel-level-test2} and Table~\ref{object-level-test2}
The mean IoU in Table~\ref{pixel-level-test2}, and \mapf{} and \mapfn{} in Table~\ref{object-level-test2} estimated on only four classes as there are no asbestos roofs in \dtesttwo. Also, there are only two buildings of concrete found in \dtesttwo{} and these two buildings were not identified from any of the experimental models, so results for the concrete class were not added to both tables.

\begin{table}[!htbp]
  \centering
  \caption{Pixel-level accuracies on \dtestone{}. IoU refers to the IoU 
   computed on the binary outputs, where  the predictions of multi-class models were binarized.  \iouf\ refers to the macro average of the IoUs for the individual classes. The subscript \emph{Multi} indicates the end-to-end setting.}
  \begin{tabular}{@{}lrrrrrrr@{}}
    \toprule
    & \multicolumn{5}{c}{IoU-Score of each class}                   \\
    \cmidrule(lr){2-6}
    Model Name  & \multicolumn{1}{c}{\thead{Metal \\ Sheet}} & \multicolumn{1}{c}{Thatch} & \multicolumn{1}{c}{Asbestos} & \multicolumn{1}{c}{Concrete} & \multicolumn{1}{c}{\thead{No \\ Roof}} & \multicolumn{1}{c}{\iouf}  & \multicolumn{1}{c}{IoU} \\
    \midrule

    YOLOv8 & \ccell{0.807}{0.003} & \ccell{0.852}{0.038} &  \ccell{0.450}{0.023} & \ccell{0.250}{0.027} & \ccell{0.480}{0.034}  & \ccell{0.568}{0.015} &  \ccell{0.866}{0.012}  \\
    \addlinespace[2pt]

    DINOv2 & \ccell{0.804}{0.003} & \ccell{0.854}{0.003} &  \ccell{0.349}{0.010} & \ccell{0.196}{0.004} & \ccell{0.608}{0.013}  & \ccell{0.562}{0.003} &  \ccell{0.883}{0.002}  \\
    \addlinespace[2pt]

    \dinovh & \ccell{0.810}{0.003} & \ccell{0.867}{0.001} &  \ccell{0.348}{0.009} & \ccell{0.188}{0.015} & \ccell{0.613}{0.005}  & \ccell{0.565}{0.004} &  \ccell{0.884}{0.001}  \\
    \addlinespace[2pt]
    
    U-Net & \ccell{0.813}{0.009} & \ccell{0.881}{0.002} &  \ccell{0.408}{0.012} & \ccell{0.171}{0.021} & \ccell{0.577}{0.073}  & \ccell{0.570}{0.016} &  \bcell{0.895}{0.003} \\ \addlinespace[2pt]
    
    \uneth  & \ccell{0.824}{0.005} & \ccell{0.879}{0.010} &  \ccell{0.384}{0.042} & \ccell{0.174}{0.010} & \ccell{0.623}{0.028}  & \ccell{ 0.577}{0.009} &  \bcell{0.895}{0.002} \\ \addlinespace[2pt]

    \unete   & \ccell{0.824}{0.006} & \bcell{0.887}{0.002} &  \ccell{0.424}{0.055} & \ccell{0.160}{0.026} & \ccell{0.591}{0.057}  & \ccell{0.577}{0.011} &  \ccell{0.888}{0.009} \\ \addlinespace[2pt]

     \midrule

     \yolom & \ccell{0.750}{0.030} & \ccell{0.824}{0.004} &  \ccell{0.405}{0.021} & \ccell{0.223}{0.059} & \ccell{0.549}{0.026}  & \ccell{0.550}{0.017} &  \ccell{0.824}{0.023}  \\

     \dinovs & \ccell{0.821}{0.003} & \ccell{0.862}{0.002} &  \ccell{0.490}{0.026} & \ccell{0.682}{0.031} & \ccell{0.640}{0.014}  & \ccell{0.699}{0.012} &  \ccell{0.880}{0.000}  \\
    \addlinespace[2pt]

    \dinovsh & \ccell{0.839}{0.002} & \ccell{0.870}{0.002} &  \bcell{0.542}{0.009} & \bcell{0.773}{0.009} & \ccell{0.649}{0.015}  & \bcell{0.734}{0.006} &  \ccell{0.885}{0.001}  \\
    \addlinespace[2pt]
    
    \unetm & \ccell{0.819}{0.012} & \ccell{0.880}{0.004} &  \ccell{0.514}{0.025} & \ccell{0.306}{0.091} & \bcell{0.650}{0.029}  & \ccell{0.634}{0.024} &  \ccell{0.879}{0.012}  \\
    \addlinespace[2pt]

    \unethm & \bcell{0.827}{0.011} & \bcell{0.887}{0.002} &  \ccell{0.511}{0.044} & \ccell{0.290}{0.105} & \ccell{0.636}{0.013}  & \ccell{0.630}{0.026} &  \ccell{0.889}{0.009}  \\
    \addlinespace[2pt]
      
    \bottomrule
  \end{tabular}
  \label{pixel-level}
\end{table}

\begin{table}[!htbp]
  \centering
  \caption{Object-level accuracy on \dtestone{}. We report the AP for each roof type, and $\mapf$ and  $\mapfn$ are  macro averages over the roof types.
  The rightmost three columns give the results when we discard the roof type information and just consider building detection.   The \tps{} columns count true positives, where \tps{} are the number of objects that overlap with ground truth objects with a minimum IoU of $0.5$. The total number of ground truth objects in the \dtestone{} is $2527$.}
  \resizebox{\textwidth}{!}{%
  \begin{tabular}{@{}lrrrrrrrrrrr@{}}
    \toprule
    & \multicolumn{5}{c}{\apf{} of each class}  & \multicolumn{3}{c}{average over classes} & \multicolumn{3}{c}{ignoring roof type} \\
    \cmidrule(lr){2-6} \cmidrule(lr){7-9} \cmidrule(lr){10-12}
    Model Name  & \thead{Metal \\ Sheet} & Thatch & Asbestos & Concrete & \thead{No \\ Roof} & \mapf  & \mapfn & \tps & \apf & \apfn & \tps \\
    \midrule
    
    YOLOv8 & \ccell{0.841}{0.003} & \ccell{0.945}{0.008} & \ccell{0.505}{0.032} & \ccell{0.542}{0.055} & \ccell{0.661}{0.026} & \ccell{0.698}{0.018} & \ccell{0.548}{0.010} & \ccell{2262.2}{7.386} & \bcell{0.941}{ 0.003} & \ccell{0.798}{0.002} & \bcell{2405.0}{5.514} \\ 
    \addlinespace[2pt]
    
    DINOv2  & \ccell{0.807}{0.005} & \ccell{0.885}{0.006} & \ccell{0.470}{0.011} & \ccell{0.579}{0.025} & \ccell{0.673}{0.015} & \ccell{0.683}{0.008} & \ccell{0.531}{0.005} & \ccell{2135.6}{7.761} & \ccell{0.882}{ 0.004} & \ccell{0.733}{0.003} & \ccell{2261.6}{9.351} \\
    \addlinespace[2pt]

    \dinovh  & \ccell{0.852}{0.002} & \ccell{0.940}{0.001} & \ccell{0.517}{0.016} & \ccell{0.600}{0.028} & \ccell{0.715}{0.006} & \ccell{0.725}{0.004} & \ccell{0.573}{0.008} & \ccell{2238.4}{5.238} & \ccell{0.930}{0.005} & \ccell{0.781}{0.005} & \ccell{2376.2}{7.194} \\
    \addlinespace[2pt]
    
    U-Net & \ccell{0.826}{0.005} & \ccell{0.924}{0.006} & \ccell{0.499}{0.016} & \ccell{0.511}{0.042} & \ccell{0.679}{0.015} & \ccell{0.688}{0.014} & \ccell{0.578}{0.014} & \ccell{2191.2}{11.25} & \ccell{0.910}{ 0.005} & \ccell{0.797}{0.003} & \ccell{2323.0}{6.033} \\ \addlinespace[2pt]

    \uneth  & \ccell{0.855}{0.005} & \bcell{0.946}{0.005} & \ccell{0.545}{0.019} & \bcell{0.596}{0.049} & \ccell{0.707}{0.012} & \bcell{0.730}{0.011} & \bcell{0.614}{0.007} & \ccell{2249.4}{4.128} & \ccell{0.935}{0.001} & \bcell{0.819}{0.003} & \ccell{2383.6}{5.314} \\ \addlinespace[2pt]

    \unete  & \ccell{0.851}{0.006} & \ccell{0.943}{0.004} & \bcell{0.551}{0.011} & \ccell{0.587}{0.049} & \ccell{0.687}{0.022} & \ccell{0.724}{0.007} & \ccell{0.606}{0.005} & \ccell{2243.2}{3.487} & \ccell{0.929}{ 0.004} & \ccell{0.818}{0.002} & \ccell{2374.4}{5.783} \\ \addlinespace[2pt]

    \midrule

    \yolom & \ccell{0.849}{0.006} & \ccell{0.923}{0.007} & \ccell{0.467}{0.035} & \ccell{0.070}{0.027} & \ccell{0.676}{0.020} & \ccell{0.597}{0.007} & \ccell{0.481}{0.003} & \bcell{2195.6}{15.383} & \ccell{0.910}{ 0.005} & \ccell{0.751}{0.007} & \ccell{2328.2}{9.988} \\
    
    \dinovs & \ccell{0.869}{0.003} & \ccell{0.923}{0.005} & \ccell{0.474}{0.043} & \ccell{0.508}{0.100} & \ccell{0.669}{0.033} & \ccell{0.689}{0.025} & \ccell{0.512}{0.021} & \ccell{2231.6}{2.332} & \ccell{0.899}{ 0.003} & \ccell{0.733}{0.004} & \ccell{2311.4}{2.728} \\ \addlinespace[2pt]

    \dinovsh & \bcell{0.888}{0.004} & \ccell{0.937}{0.004} & \ccell{0.536}{0.018} & \ccell{0.504}{0.056} & \ccell{0.647}{0.028} & \ccell{0.702}{0.013} & \ccell{0.558}{0.011} & \ccell{2270.4}{9.308} & \ccell{0.918}{ 0.003} & \ccell{0.766}{0.002} & \ccell{2366.0}{7.266} \\ \addlinespace[2pt]
    
    \unetm   & \ccell{0.883}{0.009} & \ccell{0.940}{0.010} & \ccell{0.531}{0.022} & \ccell{0.498}{0.051} & \bcell{0.728}{0.040} & \ccell{0.716}{0.018} & \ccell{0.603}{0.011} & \ccell{2262.4}{6.119} & \ccell{0.924}{ 0.004} & \ccell{0.797}{0.007} & \ccell{2358.4}{9.091} \\ \addlinespace[2pt]

    \unethm   & \ccell{0.864}{0.001} & \ccell{0.918}{0.006} & \ccell{0.533}{0.024} & \ccell{0.537}{0.048} & \ccell{0.702}{0.018} & \ccell{0.711}{0.010} & \ccell{0.609}{0.012} & \ccell{2216.2}{7.305} & \ccell{0.903}{ 0.004} & \ccell{0.786}{0.003} & \ccell{2308.0}{5.044} \\ \addlinespace[2pt]

    \bottomrule
  \end{tabular}}
  \label{object-level}
\end{table}

\begin{table}[!htbp]
  \centering
  \caption{Pixel-level accuracies on \dtesttwo{}. IoU refers to the IoU 
   computed on the binary outputs, where the predictions of multi-class models were binarized.  \iouf\ refers to the macro average of the IoUs for the individual classes. The subscript \emph{Multi} indicates the end-to-end setting.}
  \begin{tabular}{@{}lrrrrr@{}}
    \toprule
    & \multicolumn{3}{c}{IoU-Score of each class}                   \\
    \cmidrule(lr){2-4}
    Model Name  & \multicolumn{1}{c}{\thead{Metal \\ Sheet}} & \multicolumn{1}{c}{Thatch} & \multicolumn{1}{c}{\thead{No \\ Roof}} & \multicolumn{1}{c}{\thead{IoU \\ (Mean)}}  & \multicolumn{1}{c}{\thead{IoU \\ (Binary)}} \\
    \midrule
    
    YOLOv8 & \ccell{0.888}{0.003} & \ccell{0.879}{0.003} & \ccell{0.516}{0.011}  & \ccell{0.761}{0.006} &  \ccell{0.896}{0.002}  \\ \addlinespace[2pt]  

    DINOv2 & \ccell{0.867}{0.014} & \ccell{0.853}{0.004} & \ccell{0.523}{0.022}  & \ccell{0.747}{0.011} &  \ccell{0.905}{0.000}  \\ \addlinespace[2pt]
    
    \dinovh & \ccell{0.891}{0.003} & \ccell{0.880}{0.002} & \ccell{0.560}{0.018}  & \ccell{0.777}{0.007} &  \ccell{0.905}{0.001}  \\ \addlinespace[2pt]  
    
    U-Net & \ccell{0.896}{0.005} & \ccell{0.883}{0.005} & \ccell{0.463}{0.017}  & \ccell{0.748}{0.007} &  \ccell{0.909}{0.001} \\ \addlinespace[2pt]
    
    \uneth  & \ccell{0.905}{0.002} & \bcell{0.895}{0.003} & \ccell{0.493}{0.018}  & \ccell{0.764}{0.006} &  \bcell{0.911}{0.002} \\ \addlinespace[2pt]

    \unete   & \ccell{0.900}{0.008} & \ccell{0.889}{0.002} & \ccell{0.452}{0.031}  & \ccell{0.747}{0.009} &  \ccell{0.902}{0.003} \\ \addlinespace[2pt]

    \midrule

    \yolom  & \ccell{0.890}{0.006} & \ccell{0.860}{0.006} & \ccell{0.606}{0.019}  & \ccell{0.785}{0.006} &  \ccell{0.885}{0.002}  \\ 
    \addlinespace[2pt]

    \dinovs & \ccell{0.905}{0.002} & \ccell{0.875}{0.004} & \bcell{0.674}{0.018}  & \ccell{0.818}{0.005} &  \ccell{0.899}{0.002}  \\
    
    \dinovsh & \ccell{0.912}{0.001} & \ccell{0.881}{0.002} & \ccell{0.663}{0.017}  & \bcell{0.875}{0.010} &  \ccell{0.902}{0.001}  \\
    
    \unetm & \ccell{0.913}{0.005} & \ccell{0.884}{0.003} & \ccell{0.617}{0.061}  & \ccell{0.805}{0.020} &  \ccell{0.903}{0.002}  \\
    \addlinespace[2pt]
    
    \unethm & \bcell{0.921}{0.001} & \ccell{0.888}{0.002} & \ccell{0.613}{0.033}  & \ccell{0.807}{0.011} &  \ccell{0.909}{0.002}  \\
    \addlinespace[2pt]
    
    \bottomrule
  \end{tabular}
  \label{pixel-level-test2}
\end{table}

\begin{table}[!htbp]
  \centering
  \caption{Object-level accuracies on \dtesttwo{}. We report the AP for each roof type, and $\mapf$ and  $\mapfn$ are  macro averages over the classes.
  The rightmost three columns give the results when we discard the roof type information and just consider building detection.
  \tps{} are the number of objects that overlap with ground truth objects with a minimum IoU of $0.5$. The total number of ground truth objects in the \dtesttwo{} is $1541$.}
  \resizebox{\textwidth}{!}{%
  \begin{tabular}{@{}lrrrrrrrrr@{}}
    \toprule
    & \multicolumn{3}{c}{\apf of each class}  & \multicolumn{3}{c}{Objects with Classes} & \multicolumn{3}{c}{Only Building Objects} \\
    \cmidrule(lr){2-4} \cmidrule(lr){5-7} \cmidrule(lr){8-10}
    Model Name  & \multicolumn{1}{c}{\thead{Metal \\ Sheet}} & \multicolumn{1}{c}{Thatch} & \multicolumn{1}{c}{\thead{No \\ Roof}} & \multicolumn{1}{c}{\mapf}  & \multicolumn{1}{c}{\mapfn} & \multicolumn{1}{c}{\tps} & \multicolumn{1}{c}{\apf} & \multicolumn{1}{c}{\apfn} & \multicolumn{1}{c}{\tps} \\
    \midrule
    
    YOLOv8   & \ccell{0.928}{0.001} & \bcell{0.947}{0.000} & \ccell{0.661}{0.023} & \ccell{0.846}{0.008} & \ccell{0.428}{0.002} & \ccell{1447.2}{4.534} & \bcell{0.963}{ 0.005} & \ccell{0.838}{0.002} & \bcell{1493.8}{3.826} \\ 
     \addlinespace[2pt]

     \dinovs   & \ccell{0.898}{0.004} & \ccell{0.885}{0.009} & \ccell{0.635}{0.024} & \ccell{0.484}{0.005} & \ccell{0.393}{0.005} & \ccell{1381.6}{7.172} & \ccell{0.919}{ 0.005} & \ccell{0.786}{0.006} & \ccell{1428.8}{7.305} \\ 
     \addlinespace[2pt]

    \dinovsh   & \ccell{0.932}{0.002} & \ccell{0.942}{0.005} & \ccell{0.681}{0.02
    0} & \ccell{0.852}{0.007} & \ccell{0.423}{0.003} & \ccell{1441.8}{4.400} & \ccell{0.956}{ 0.001} & \ccell{0.828}{0.003} & \ccell{1486.2}{1.939} \\ 
     \addlinespace[2pt]
     
    U-Net  & \ccell{0.915}{0.006} & \ccell{0.921}{0.006} & \ccell{0.520}{0.027} & \ccell{0.590}{0.006} & \ccell{0.407}{0.004} & \ccell{1399.4}{4.758} & \ccell{0.929}{ 0.000} & \ccell{0.836}{0.002} & \ccell{1438.4}{4.499} \\ \addlinespace[2pt]
    
    \uneth  & \ccell{0.932}{0.003} & \ccell{0.946}{0.004} & \ccell{0.559}{0.027} & \ccell{0.812}{0.008} & \ccell{0.528}{0.003} & \ccell{1429.0}{6.229} & \ccell{0.947}{0.004} & \bcell{0.858}{0.004} & \ccell{1468.6}{6.499} \\ \addlinespace[2pt]
     
    \unete  & \ccell{0.935}{0.001} & \ccell{0.940}{0.004} & \ccell{0.509}{0.022} & \ccell{0.795}{0.008} & \ccell{0.518}{0.005} & \ccell{1421.0}{3.688} & \ccell{0.939}{ 0.000} & \ccell{0.851}{0.004} & \ccell{1458.8}{4.118} \\ \addlinespace[2pt]

    \midrule

    \yolom   & \ccell{0.949}{0.004} & \ccell{0.934}{0.007} & \ccell{0.664}{0.044} & \ccell{0.849}{0.015} & \ccell{0.423}{0.008} & \ccell{1446.0}{4.899} & \ccell{0.948}{ 0.003} & \ccell{0.808}{0.005} & \ccell{1477.2}{3.655} \\ 
    \addlinespace[2pt]

    \dinovs & \ccell{0.955}{0.004} & \ccell{0.935}{0.007} & \bcell{0.749}{0.030} & \bcell{0.880}{0.011} & \ccell{0.539}{0.005} & \bcell{1454.4}{3.878} & \ccell{0.946}{ 0.001} & \ccell{0.801}{0.003} & \ccell{1468.8}{2.926} \\

    \dinovsh & \bcell{0.956}{0.001} & \ccell{0.943}{0.005} & \ccell{0.727}{0.026} & \ccell{0.875}{0.010} & \ccell{0.521}{0.047} & \ccell{1460.8}{3.868} & \ccell{0.950}{ 0.005} & \ccell{0.820}{0.002} & \ccell{1478.8}{3.311} \\
    
    \unetm   & \bcell{0.956}{0.004} & \ccell{0.926}{0.008} & \ccell{0.651}{0.107} & \ccell{0.844}{0.039} & \bcell{0.548}{0.017} & \ccell{1439.0}{16.358} & \ccell{0.943}{ 0.010} & \ccell{0.838}{0.006} & \ccell{1463.6}{13.063} \\ \addlinespace[2pt]
    
    \unethm   & \ccell{0.951}{0.004} & \ccell{0.920}{0.006} & \ccell{0.632}{0.029} & \ccell{0.834}{0.010} & \ccell{0.458}{0.005} & \ccell{1426.8}{5.418} & \ccell{0.947}{ 0.001} & \ccell{0.854}{0.004} & \ccell{1472.6}{2.417} \\ \addlinespace[2pt]
    \bottomrule
  \end{tabular}}
  \label{object-level-test2}
\end{table}

\subsection{Performance of Different Classifiers}\label{cls-results}\label{app:classifier}
In the two-stage approach, we used a classifier based on DINOv2 features, as described in Section~\ref{sec:twostage} and illustrated in Figure~\ref{dinovc}.
The input representation was fixed and was processed by standard classification algorithms. We compared linear probing based on logistic regression with $L_2$-regularization
and k-nearest neighbours (kNN) classification trained on our data.
For evaluating the  classifiers and tuning their hyperparameters, we combined the training and validation data and performed  10-fold cross-validation (CV) with F1-score as performance metric. 
The best CV results gave logistic regression with $L_2$-regularization, and this model was used for all subsequent two-stage experiments, see Table~\ref{cls-results}.
\begin{table}[ht!] 
  \centering
  \caption{Cross-validation accuracies on combined training and validation data of k-nearest neighbour classification (kNN) and logistic regression applied to the DINOv2 features. The baseline is the architecture depicted in Figure~\ref{dinovc}, \emph{w/o mask} refers to omitting the masking and averaging the DINOv2 features across the whole input patch, and \emph{w/o upsampling} did not upsample the DINOv2 features but downsampled the  building mask instead.}
  \begin{tabular}{@{}lcccccc@{}}
    \toprule
    & \multicolumn{6}{c@{}}{F1-Score}                   \\
    \cmidrule(r){2-7}
    & \multicolumn{2}{c}{baseline}  & \multicolumn{2}{c}{w/o mask} & \multicolumn{2}{c}{w/o upsampling} \\
    \cmidrule(lr){2-3} \cmidrule(lr){4-5} \cmidrule(lr){6-7}
    
    Classifier & Mean & Std & Mean & Std & Mean  & Std\\
    \midrule
    Logistic Regression &  \textbf{0.770} & 0.063  & 0.573 & 0.077 & 0.768 & 0.067 \\
    \addlinespace[4pt]
    kNN & 0.734 & 0.045  & 0.389 & 0.029 & 0.733 & 0.051  \\ 
    \bottomrule
  \end{tabular}
  \label{classifier}
\end{table}

We also performed an ablation study to show the importance of the masking and the upsampling in our architecture shown in Figure~\ref{dinovc}. The results are also depicted in  Table~\ref{cls-results}. When we omitted the masking and considered all features, the results got considerably worse.
If we omitted the upsampling of the DINOv2 output and downsampled the masks instead, the performance also slightly dropped.

\end{document}

%% file: figures/Figure_d_arXiv.tex
\begingroup%
  \makeatletter%
  \providecommand\color[2][]{%
    \errmessage{(Inkscape) Color is used for the text in Inkscape, but the package 'color.sty' is not loaded}%
    \renewcommand\color[2][]{}%
  }%
  \providecommand\transparent[1]{%
    \errmessage{(Inkscape) Transparency is used (non-zero) for the text in Inkscape, but the package 'transparent.sty' is not loaded}%
    \renewcommand\transparent[1]{}%
  }%
  \providecommand\rotatebox[2]{#2}%
  \newcommand*\fsize{\dimexpr\f@size pt\relax}%
  \newcommand*\lineheight[1]{\fontsize{\fsize}{#1\fsize}\selectfont}%
  \ifx\svgwidth\undefined%
    \setlength{\unitlength}{425bp}%
    \ifx\svgscale\undefined%
      \relax%
    \else%
      \setlength{\unitlength}{\unitlength * \real{\svgscale}}%
    \fi%
  \else%
    \setlength{\unitlength}{\svgwidth}%
  \fi%
  \global\let\svgwidth\undefined%
  \global\let\svgscale\undefined%
  \makeatother%
  \rmfamily
  \begin{picture}(1,0.52042835)%
    \lineheight{1}%
    \setlength\tabcolsep{0pt}%
    \put(0,0){\includegraphics[width=\unitlength,page=1]{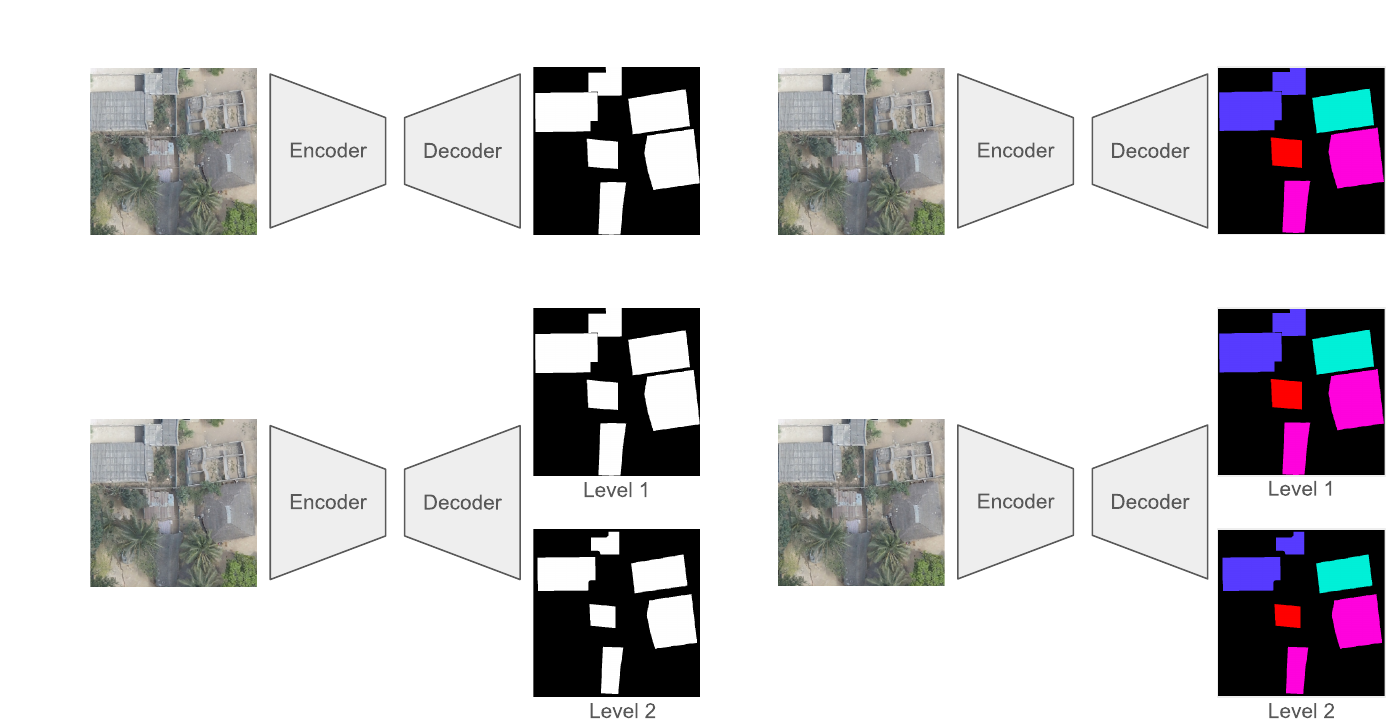}}%
    \put(0.01826876,0.41051055){\color[rgb]{0,0,0}\rotatebox{90}{\makebox(0,0)[t]{\lineheight{1.25}\smash{\begin{tabular}[t]{c}Baseline\end{tabular}}}}}%
    \put(0.55890406,0.50215933){\color[rgb]{0,0,0}\makebox(0,0)[lt]{\lineheight{1.25}\smash{\begin{tabular}[t]{l}Roof Material Segmentation\\\end{tabular}}}}%
    \put(0.06286334,0.50215959){\color[rgb]{0,0,0}\makebox(0,0)[lt]{\lineheight{1.25}\smash{\begin{tabular}[t]{l}Building Segmentation\\\end{tabular}}}}%
    \put(0.01900966,0.16005851){\color[rgb]{0,0,0}\rotatebox{90}{\makebox(0,0)[t]{\lineheight{1.25}\smash{\begin{tabular}[t]{c}Deep Ordinal Watershed \\(DOW)\\\end{tabular}}}}}%
    \put(0.193,0.33){\color[rgb]{0,0,0}\makebox(0,0)[lt]{\lineheight{1.25}\smash{\begin{tabular}[t]{l}\scriptsize\{UNet, DINOv2\}\\\end{tabular}}}}%
    \put(0.69,0.33){\color[rgb]{0,0,0}\makebox(0,0)[lt]{\lineheight{1.25}\smash{\begin{tabular}[t]{l}\scriptsize\{UNet, DINOv2\}$_\text{Multi}$\\\end{tabular}}}}%
    \put(0.193,0.079){\color[rgb]{0,0,0}\makebox(0,0)[lt]{\lineheight{1.25}\smash{\begin{tabular}[t]{l}\scriptsize\{UNet, DINOv2\}$_\text{DOW}$\\\\\end{tabular}}}}%
    \put(0.69,0.079){\color[rgb]{0,0,0}\makebox(0,0)[lt]{\lineheight{1.25}\smash{\begin{tabular}[t]{l}\scriptsize\{UNet, DINOv2\}$_\text{DOW-Multi}$\\\\\end{tabular}}}}%
  \end{picture}%
\endgroup%